\documentclass[conference]{IEEEtran}
\IEEEoverridecommandlockouts
\usepackage{cite}
\usepackage{amsmath,amssymb,amsfonts}
\usepackage{algorithmic}
\usepackage{graphicx}
\usepackage{textcomp}
\usepackage{xcolor}
\def\BibTeX{{\rm B\kern-.05em{\sc i\kern-.025em b}\kern-.08em
    T\kern-.1667em\lower.7ex\hbox{E}\kern-.125emX}}

\usepackage{todonotes}
\usepackage{multirow}
\usepackage{authblk}

\def\ie{\emph{i.e.~}}
\def\eg{\emph{e.g.~}}
\def\cf{\emph{cf.~}}
\def\wrt{wrt.~}

\newcommand{\lrpz}[0]{\emph{LRP$_{z}$}}
\newcommand{\LRPZ}[0]{\textbf{\lrpz}}

\newcommand{\lrpeps}[0]{\emph{LRP$_{\varepsilon}$}}
\newcommand{\LRPEPS}[0]{\textbf{\lrpeps}}

\newcommand{\lrpalphabeta}[0]{\emph{LRP$_{\alpha\beta}$}}
\newcommand{\LRPALPHABETA}[0]{\textbf{\lrpalphabeta}}

\newcommand{\lrpflat}[0]{\emph{LRP$_\flat$}}
\newcommand{\lrpzb}[0]{\emph{DTD$_{z^B}$}}

\newcommand{\lrpcnn}[0]{\emph{LRP$_{CMP}$}}

\newcommand{\LRPCMP}[0]{\textbf{\lrpcnn}}

\newcommand{\lrpalphaonebetazero}[0]{\emph{LRP$_{\alpha1}$}}
\newcommand{\LRPALPHAONEBETAZERO}[0]{\textbf{\lrpalphaonebetazero}}

\newcommand{\lrpalphatwobetaone}[0]{\emph{LRP$_{\alpha2}$}}
\newcommand{\LRPALPHATWOBETAONE}[0]{\textbf{\lrpalphatwobetaone}}

\newcommand{\lrpcnnalphaonebetazero}[0]{\emph{LRP$_{CMP:{\alpha1}}$}}
\newcommand{\LRPCNNALPHAONEBETAZERO}[0]{\textbf{\lrpcnnalphaonebetazero}}

\newcommand{\lrpcnnalphatwobetaone}[0]{\emph{LRP$_{CMP:{\alpha2}}$}}
\newcommand{\LRPCNNALPHATWOBETAONE}[0]{\textbf{\lrpcnnalphatwobetaone}}

\newcommand{\lrpcnnalphaonebetazeroflat}[0]{\emph{LRP$_{CMP:{\alpha1+\flat}}$}}
\newcommand{\LRPCNNALPHAONEBETAZEROFLAT}[0]{\textbf{\lrpcnnalphaonebetazeroflat}}

\newcommand{\lrpcnnalphatwobetaoneflat}[0]{\emph{LRP$_{CMP:{\alpha2+\flat}}$}}
\newcommand{\LRPCNNALPHATWOBETAONEFLAT}[0]{\textbf{\lrpcnnalphatwobetaoneflat}}

\title{Towards Best Practice in Explaining Neural Network Decisions with LRP\\
\thanks{
    This work was partly supported by the German Ministry for Education and Research as
    BIFOLD
    (refs. 01IS18025A and 01IS18037A)
    and TraMeExCo (ref. 01IS18056A).
    A. Binder is grateful for the support by the Ministry of Education of Singapore (MoE) Tier 2 grant MOE2016-T2-2-154.
    This publication only reflects the authors views.
    Funding agencies are not liable for any use that may be made of the information contained herein.}
}

\author[1]{Maximilian Kohlbrenner} 
\author[2]{Alexander Bauer} 
\author[2]{Shinichi Nakajima}
\author[3]{Alexander Binder}
\author[1,$\ast$]{\\Wojciech Samek}
\author[1,$\ast$]{Sebastian Lapuschkin}

\affil[1]{Dept.~of Video Coding and Analytics,
          Fraunhofer Heinrich Hertz Institute, Berlin,  Germany
         }
\affil[2]{Dept.~of Electrical Engineering and Computer Science,
	  Technische Universit\"at Berlin, Berlin,  Germany
	  }
\affil[3]{ISTD Pillar,
          Singapore University of Technology and Design, Singapore, Singapore}
\affil[$\ast$]{\small\texttt{\{wojciech.samek|sebastian.lapuschkin\}@hhi.fraunhofer.de}}

\begin{document}

\maketitle
\begin{abstract}
Within the last decade, neural network based predictors have demonstrated impressive --- and at times super-human --- capabilities.
This performance is often paid for with an intransparent prediction process and thus has
sparked numerous contributions in the novel field of \emph{explainable artificial intelligence (XAI)}.
In this paper, we focus on a popular and widely used method of XAI, the \emph{Layer-wise Relevance Propagation (LRP)}.
Since its initial proposition LRP has evolved as a method, and a \emph{best practice} for applying the method has
tacitly emerged, based however on humanly observed evidence alone.
In this paper we investigate --- and for the first time \emph{quantify} --- the effect of this current best practice on feedforward neural networks in a visual object detection setting.
The results verify that the layer-dependent approach to LRP applied in recent literature better represents the model's reasoning,
and at the same time increases the object localization and class discriminativity of LRP.

\end{abstract}

\section{Introduction}
In recent years, deep neural networks (\emph{DNN}) have become the state of the art method in many different fields,
but are mainly applied as black-box predictors.
Since understanding the decisions of artificial intelligence systems is crucial in numerous scenarios and partially demanded by law\footnote{\eg via the ``right to explanation'' proclaimed in the General Data Protection Regulation of the European Union~\cite{GDPR2016,goodman2016european}},
neural network interpretability has been established as an important and active research area.
Consequently, many approaches to explaining neural network decisions have been proposed in recent years,
\eg
\cite{springenberg2014striving,kindermans2017learning,sundararajan2017axiomatic,smilkov2017smoothgrad}.
The \emph{Layer-wise Relevance Propagation} (\emph{LRP}) \cite{bach2015pixel} framework has proven successful at providing a meaningful intuition and measurable quantities describing a network's feature processing and decision making
\cite{yang2018explaining, thomas2018analyzing,lapuschkin2019unmasking}.
LRP attributes \emph{relevance scores} $R_i$ to the model inputs or intermediate neurons $i$ by decomposing a model output of interest.
The method follows the principles of \emph{relevance conservation} and \emph{proportional decomposition}.
Therefore, attributions computed with LRP maintain a strong connection to the predictor output.
While early applications of LRP administer a single decomposition rule uniformly to all layers of a model
\cite{bach2015pixel,lapuschkin2016analyzing, ancona2019gradient}, more recent work describes a trend towards assigning specific decomposition rules purposedly to layers \wrt function and position within the network
\cite{montavon2019layer, lapuschkin2019unmasking, lapuschkin2017understanding, hagele2019resolving, hui2019batchnorm}.
This trend has tacitly emerged and formulates a \emph{best practice} for applying LRP.
Under qualitative evaluation, the attribution maps resulting from this current approach seem to be more robust against the well-known effects of
shattered gradients
\cite{ancona2019gradient,balduzzi2017shattered,montavon2019layer}
and demonstrate an increased discriminativity between different target classes
\cite{montavon2019layer, lapuschkin2017understanding}
compared to the uniform application of a single rule.

However, recent literature applying LRP-rules in a layer-dependent manner
do not justify the beneficial effects of this novel variant \emph{quantitatively}, but only based on human observation.
In this paper, we design and conduct a series of experiments in order to verify whether a layer-specific application of different decomposition rules actually constitutes an improvement above earlier descriptions and applications of LRP~\cite{lapuschkin2016analyzing,samek2016evaluating}.
That is,
we measure and compare capabilities of various methods from explainable AI --- with a focus on earlier and more recent approaches to LRP ---
to \emph{precisely} localize the ground-truth objects in images via attribution of relevance scores.
Our experiments are conducted on popular computer vision data sets with ground truth object localizations, the ImageNet~\cite{ILSVRC15} and PascalVOC~\cite{pascal-voc-2012} datasets, using different neural network models.

\section{Feedforward Neural Networks and LRP }
\label{seq:lrp-sequential}

\begin{figure*}[!h]
  \begin{center}
    \includegraphics[width=\textwidth]{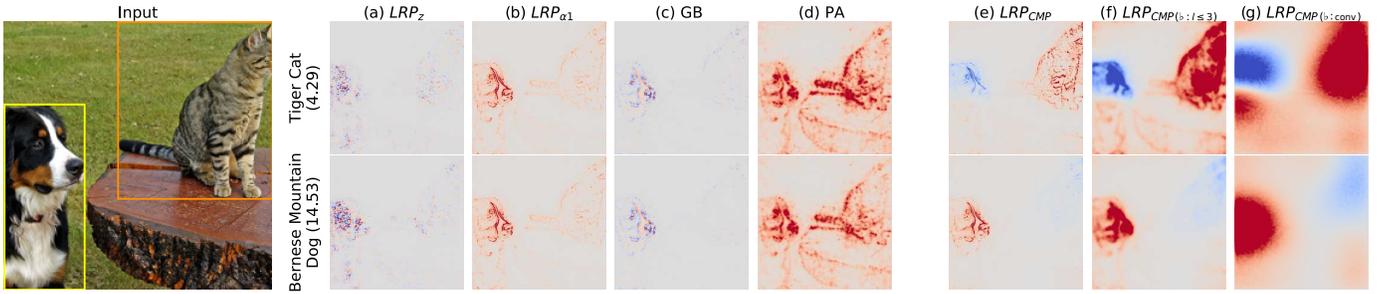}
    \caption{Different attributions for the output classes ``Tiger Cat'' and ``Bernese Mountain Dog'' using the VGG-16 model.
    Network output strengths (logit) of the respective classes is given in parentheses.
    Network-widely applied rules in
    \emph{(a)} - \emph{(d)} (\lrpz, \lrpalphabeta, Guided Backprop and Pattern Attribution respectively),
    are not, or hardly class discriminative.
    An application of \lrpz~to every layer shows the effect of gradient shattering.
    Variants of \LRPCMP~implementing a composite strategy of LRP rule application shown in \emph{(e)} - \emph{(g)} --- here, from left to right, the \lrpflat-rule  is not applied at all, the three lowest convolution layers, and all convolution and pooling layers --- are sensitive to class-specific information and highlight features on different levels of scale and conceptuality (\eg highlighting the fur pattern activating ``Tiger Cat'' vs highlighting the general region showing a ``Tiger Cat'').
    Attributions from \LRPCMP~visualized in red/warm colors identify image regions contributing to the prediction of the target class, while regions marked in blue/cold hues provide contradictory evidence.
    Further examples can be found in the Appendix.}

    \label{fig:ex_image_catdog}
  \end{center}
\end{figure*}

Feedforward neural networks constitute a popular architecture type, ranging from simple multi-layer perceptrons and shallower convolutional architectures
such as the LeNet-5 \cite{lecun1998gradient}
to deeper and more complex Inception \cite{szegedy2016rethinking} and VGG-like architectures \cite{simonyan2014very}.
These types of neural network commonly use ReLU non-linearities and first pass information through a stack of convolution and pooling layers, followed by several fully connected layers.
The good performance of feedforward architectures in numerous problem domains,
and the availability as pre-trained models makes them a valuable standard architecture in neural network design.

\subsection{Layer-wise Relevance Propagation}
Consequently, feedforward networks have been subject to investigations in countless contributions towards neural network interpretability, including applications of LRP~\cite{bach2015pixel,lapuschkin2016analyzing,samek2016evaluating},
which finds its mathematical foundation in \emph{Deep Taylor Decomposition (DTD)} \cite{montavon2017explaining}.

The most basic attribution rule of LRP (to which we will refer to as \lrpz) is defined as
\begin{align}
  R_i^{(l)} = \sum_j \frac{z_{ij}}{z_j} R_j^{(l+1)}
  \label{eq:lrp-basic}
\end{align}
and performs a proportional decomposition of a given upper layer relevance value $R_j^{(l+1)}$ at some layer \mbox{($l+1$)}  and neuron $j$ to obtain lower layer relevance scores $R_i^{(l)}$ for neurons $i$ at layer ($l$), \wrt to the localized preactivations $z_{ij}$ and their respective aggregations $z_j$ at the layer output.
Here, the localized preactivations $z_{ij}$ describe quantities propagated through the model during prediction time, \eg $z_{ij} = x_iw_{ij}$ and $z_j = \sum_i z_{ij}$ within a neural network layer with learned weight parameters $w_{ij}$.
Note that Eq.~\eqref{eq:lrp-basic} is conservative between layers and in general maintains an equality $\sum_i R_i^{(l)} = f(x)$ at any layer ($l$) of the model.

Further purposed LRP-rules beyond Eq.~\eqref{eq:lrp-basic} are introduced in \cite{bach2015pixel}, which can be understood as advancements thereof:

So does the \lrpeps~decomposition rule~\cite{bach2015pixel} add a signed and small constant $\varepsilon$ to the denominator in order to prevent divisions by zero and to diminish the effect of recessive (\eg weak and noisy) mappings $z_{ij}$ to the relevance decomposition.
\begin{align}
	R_i^{(l)} = \sum_j \frac{z_{ij}}{z_j + \varepsilon \cdot sign(z_j)} R_j^{(l+1)}
  \label{eq:lrp-epsilon}
\end{align}

The \lrpalphabeta-rule~\cite{bach2015pixel} performs and then merges separate decompositions for the activatory ($z_{ij}^+$) and inhibitory ($z_{ij}^-$) parts of the forward pass
\begin{align}
	R_i^{(l)} = \sum_j \big( \alpha \frac{z_{ij}^+ }{z_j^+} + \beta \frac{z_{ij}^-}{z_j^-} \big)  R_j^{(l+1)}
  \label{eq:lrp-alphabeta}
\end{align}
where
\begin{align}
  z_{ij}^+ = \begin{cases} z_{ij} &; z_{ij} > 0 \\ 0 &; \text{else} \end{cases} & & z_{ij}^- = \begin{cases} 0 &; \text{else} \\ z_{ij} &; z_{ij} < 0 \end{cases}
\end{align}
Here, the non-negative $\alpha$ parameter permits a weighting of relevance distribution towards activations and inhibitions.
The $\beta$ parameter is given implicitly s.t. $\alpha+\beta = 1$ in order to uphold conservativity of relevance between layers.
The commonly used parameter $\alpha=1$
can be derived from DTD and has been rediscovered in \emph{ExcitationBackprop}~\cite{zhang2018top}.

Later work~\cite{bach2016controlling,lapuschkin2017understanding} introduces \lrpflat\footnote{read: $\flat = $``flat'', as in the musical $\flat$.},
a decomposition rule which spreads the relevance of a neuron uniformly across all its inputs.
This rule assumes $z_{ij}=1$ and $z_j = \sum_i 1$ in Eq.~\eqref{eq:lrp-basic} \emph{only} for backpropagating given relevance scores $R_j^{(l+1)}$ to lower layers $(l)$,
and has seen application in the input layer(s) of neural networks.
The \lrpflat-rule provides invariance to the decomposition process \wrt to translations in the input domain
and effectively propagates relevance scores of higher layer neurons --- encoding ``explanations'' of more abstract concepts --- towards the input via the neurons' \emph{receptive fields}, without further transformation.
Note that the \lrpflat~decomposition rule is thus unsuitable for decomposing fully connected layers.

Earlier applications of LRP (\eg \cite{bach2015pixel,lapuschkin2016analyzing}) did use one single decomposition rule uniformly over the whole network,
which often resulted in suboptimal ``explanations'' of model behavior~\cite{montavon2019layer}.
So are network-wide applications of \lrpz~(in the following denoted as \LRPZ, in order to distinguish this specific \emph{configuration} of LRP from the \emph{rule}~\lrpz) and network-wide applications of \lrpeps~(denoted as \LRPEPS) respectively identical and highly similar to \emph{Gradient$\times$Input (G$\times$I)} in ReLU-activated DNNs~\cite{ancona2019gradient}.
\LRPZ~and \LRPEPS~demonstrate --- albeit working well for shallower convolutional models~\cite{lapuschkin2016lrp, horst2019explaining} such as the \mbox{LeNet-5}~\cite{lecun1998gradient}
or simpler fully-connected networks~\cite{StuJNM16} --- the effect of gradient shattering as overly complex attributions for deeper models~\cite{montavon2019layer,ancona2019gradient} (\cf Fig.~\ref{fig:ex_image_catdog}\emph{(a)}).
A Network-wide application of \lrpalphabeta~(denoted as \LRPALPHABETA) demonstrates robustness against gradient shattering and produces visually pleasing attribution maps,
however is lacking in class- or object discriminativity~\cite{gu2018understanding,montavon2019layer}.
By separately considering activatory and inhibitory mappings $z_{ij}$ during the decomposition process, \lrpalphabeta~tends to attribute relevance to similar sets of input features activating sequences of neurons throughout the network, regardless of the output class chosen for relevance decomposition (\cf Fig.~\ref{fig:ex_image_catdog}\emph{(b)}).
Further, \LRPALPHABETA~introduces the constraint of strictly positive layer activations~\cite{montavon2017explaining},
which is in general not guaranteed, especially at the (logit) output of a model.
A dissatisfaction of this constraint may result in a sign inversion of all backpropagated relevance scores.

\subsection{A Current Best Practice for LRP}

A recent trend among XAI researchers and practitioners employing LRP is the use of a \emph{composite strategy} of rule applications for decomposing the prediction of a neural network~\cite{lapuschkin2017understanding,montavon2019layer,lapuschkin2019unmasking,hagele2019resolving,hui2019batchnorm}. 
That is, different parts of the DNN are decomposed using purposed rules, which in combination are robust against gradient shattering while sustaining object discriminativity.
Common among these works is the utilization of \lrpeps~with  $\varepsilon \ll 1 $ (or just \lrpz) to decompose fully connected layers close to the model output,
followed by an application of \lrpalphabeta~to the underlying convolutional layers (usually with $\alpha \in \{1,2\}$).
Here, the separate decomposition of the positive and negative forward mappings complements the localized feature activation of convolutional filters activated by, and feeding into ReLUs.
A final decomposition step within the convolution layers near the input uses the \lrpflat-rule.
Most commonly this rule (or alternatively the \lrpzb-rule defined in context of Deep Taylor Decompositon~\cite{montavon2017explaining}) is applied to the input layer only.
In summary, we here describe this pattern of rule application as \LRPCMP~(for \emph{C}o\emph{MP}osite).
Fig.~\ref{fig:ex_image_catdog} provides a qualitative overview of the effect of \LRPCMP~in contrast to other parameterizations and methods, which we will further discuss in Sec.~\ref{sec:results-qualitative}.
Note that the option to apply the \lrpflat~decomposition to the first $n$ layers near the input (instead of only the first \emph{one} layer) provides control over the local and semantic scale~\cite{bach2016controlling} of the computed attributions (see Fig.~\ref{fig:ex_image_catdog}\emph{(e)}-\emph{(g)}).
Previous works profit from this option for comparing DNNs of varying depth, and differently configured convolutional stacks~\cite{lapuschkin2017understanding}, or by increasing readability of attributions maps aligned to the requirements of human investigators~\cite{hagele2019resolving}.

\section{Metric and Assumptions}

\subsection{Motivation}
The declared purpose of LRP is to precisely and quantitatively inform about the (image or intermediate) features which contribute towards or against the decision of a model \wrt to a specific predictor output~\cite{bach2015pixel}.
While the recent \LRPCMP~exhibits improved properties above previous variants of LRP \emph{by eyeballing},
an objective verification requires quantification.
The visual object detection setting, as it is described by the Pascal VOC (PVOC)~\cite{pascal-voc-2012} or ImageNet~\cite{krizhevsky2012imagenet} datasets --- both of which include object bounding box annotations --- delivers an optimal experimental setting for this purpose.

An assumed ideal model would, in such a setting, exhibit true object understanding by only predicting based on the object itself.
A good and \emph{representative} attribution method should therefore reflect that object understanding of the model closely
\ie by marking (parts of) the shown object as relevant and disregarding visual features not representing the object itself.
Similar to~\cite{lapuschkin2016analyzing}, we therefore rely on a measure based on localization of attribution scores.
In the following, we will evaluate \LRPCMP~against other methods and variants of LRP
on ImageNet using a pre-trained VGG-16 network,
and on PVOC 2007 using a pre-trained (on PVOC 2012) CaffeNet model~\cite{lapuschkin2016analyzing}.
Both models perform well on their respective task
and have been obtained from \texttt{https://modelzoo.co/}~.

\subsection{Verifying Object-centricity During Prediction}
In practice, both datasets can not be assumed to be free from contextual biases (\cf\cite{lapuschkin2019unmasking,anders2019analyzing}),
and in both settings models are trained to categorize images rather than localize objects.
Still, we (necessarily) assume that the models we use dominantly base their decision on the target object,
as opposed to the image context.

\begin{figure}[!t]
  \begin{center}
  \includegraphics[width=\linewidth]{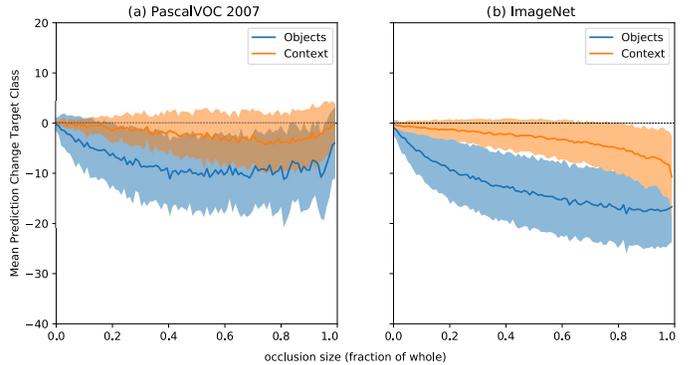}
  \end{center}
	\caption{Mean prediction changes $\Delta f(x)$ measured in the logit outputs of the true class as a function of the occluded area,
	    when occluding the pixels within (object) and without (image context) the class-specific bounding boxes
            on PVOC 2007 (\emph{left}) and ImageNet (\emph{right}). Lower values indicate a stronger reaction of the model.
            Shaded areas show the standard deviation.
          }
  \label{fig:bbx_occlusion}
\end{figure}

We verify our hypothesis in Fig.~\ref{fig:bbx_occlusion},
by showing for both models and datasets the reaction of the corresponding predictor $f$ to the occlusion of the object area vs. the occlusion of the image background.
That is, for each image $x$ of the respective dataset, we leverage the available bounding box annotations and compute partially occluded versions $x'$ where either the object area or class-specific image background (\ie the non-object area) are replaced with mean color values per corresponding pixel and dataset.
We then measure the $\Delta f(x) = f(x') - f(x)$ for the ground truth label(s) of $x$ based on the network's logit outputs, and plot this value as a function of relative bounding box size.
Fig.~\ref{fig:bbx_occlusion} shows the average values and standard deviation for $\Delta f(x)$ per bounding box size (discretized into 100 uniform bins) when replacing either the object (area within the bounding box) or the context (rest of the image).

Occluding the object area consistently leads to a sharper decrease in the output for the specific class.
The trend is especially evident for smaller objects.
This supports our claim that the networks base their decision mainly on the object itself.

\subsection{Attribution Localization as a Quantitative Measure}
This gives us a performance criterion for attribution methods in object detection and classification.
In order to track the fraction of the total amount of relevance that is attributed to the object,
we use the inside-total relevance ratio $\mu$ without, and a weighted variant $\mu_w$ within consideration of the object size:
\begin{align}
\mu =  \frac{R_{\text{in}}}{R_{\text{tot}}} && \mu_w = \mu \cdot \frac{S_{\text{tot}}}{S_{\text{in}}}
\label{eq:mu}
\end{align}
While conceptually similar to the inside-outside ratio used in \cite{lapuschkin2016analyzing},
$\mu$ and $\mu_w$ avoid numerical issues in edge cases \wrt bounding box size.
Here, $R_{\text{in}}$ is the sum of positive relevance in the bounding box,
$R_{\text{tot}}$ the total sum of positive relevance in the image and
$S_{\text{in}}$ and $S_{\text{tot}}$ are the size of the bounding box and the image respectively, in pixels.
The subscript $w$ signals the addition of a normalization factor in $\mu_w$ considering the size of image and object.

Correctly locating small objects is more difficult than locating image-sized objects.
Since the ratio $S_{\text{tot}}/S_{\text{in}}$ is always greater than or equal to 1 and increases for smaller objects,
$\mu_w$ puts additional emphasis on measuring the outcome for small bounding box sizes.
In both cases, higher values indicate larger fractions of relevance attributed to the object area (and not background),
and therefore are the desirable outcome.

\section{Experiments and Results}

\subsection{Experimental Setup}
\label{sec:setup}

We perform our experiments on both the ImageNet and the PVOC 2007 datasets,
since both collections provide large numbers of ground truth object bounding boxes.

For PVOC, we compute attribution maps for all samples (approx.~$10.000$) from PVOC 2007, using a model which has been pre-trained on the multi label setting of PVOC 2012~\cite{pascal-voc-2012,lapuschkin2016analyzing}.
The respective model performs with a mean AP of $72.12$ on PVOC 2007.
Since PVOC describes a multi label setting, multiple classes can be present in the same image.
We therefore evaluate $\mu$ and $\mu_w$ once for each unique existing pair of $\{$ class $\times$ sample $\}$,
yielding approximately $15.000$ measurements.
Images with a higher number of (smaller) bounding boxes thus effectively have a stronger impact on the results than images with larger (and fewer), image-filling objects, while at the same time describing a \emph{more difficult} setting.
Many of the objects shown in PVOC images are not centered.
In order to use all available object information in our evaluation, we rescale the input images to the network's desired input shape, avoiding the (partial) cropping of objects.

On ImageNet~\cite{ILSVRC15} (2012 version), bounding box information does only exist for the $50.000$ validation samples (displaying one class per image) and can be downloaded from the official website\footnote{\texttt{http://www.image-net.org/challenges/LSVRC/2012}}.
We evaluate a pre-trained VGG-16 model from the keras model zoo,
obtained via the iNNvestigate~\cite{alber2019innvestigate} toolbox.
The model performs with a $90.1\%$ top-5 accuracy on the ImageNet test set.
For all images the shortest side is rescaled to fit the model input and the longest side is center-cropped to obtain a quadratic input shape.
Bounding box information is adjusted correspondingly.

For computing attribution maps, we make use of existing XAI software packages, depending on the models' formats.
That is, for the VGG-16 model we use the Keras~\cite{chollet2015keras} and Tensorflow~\cite{abadi2015tensorflow} based iNNvestigate~\cite{alber2019innvestigate} toolbox.
For the PVOC data and the CaffeNet architecture, we compute attributions using the Caffe~\cite{jia2014caffe} based LRP~Toolbox~\cite{lapuschkin2016lrp}.

Both XAI packages support the same functionality regarding LRP, yet differ in the provided selection of other attribution methods.
Our study, however, shall be focussed on the beneficial or detrimental effects between the variants of LRP used in literature.

We compute attribution maps
and values for $\mu$ and $\mu_w$
for both models and
different variants of LRP: \LRPZ,
\LRPALPHABETA~(both for $\alpha=1$ and $\alpha=2$),
and
several parameterizations of \LRPCMP.
For the latter we distinguish parameter choices for $\alpha$ in a subscript when discussing quantitative results in Sec.~\ref{sec:results-quantitative}.
Additionally, in case \lrpflat~is applied to the input layer, we add ``$+\flat$'' to the subscript, \eg as ``\LRPCNNALPHAONEBETAZEROFLAT''.

We complement the results with
Guided Backprop (GB)~\cite{springenberg2014striving} and
for ImageNet with
Pattern Attribution (PA)~\cite{kindermans2017learning} only available in iNNvestigate.
On both datasets, we evaluate attributions for the ground truth class labels, independent of the network prediction.

%
%
%
%
%
%

\subsection{Qualitative Observations}
\label{sec:results-qualitative}

Fig.~\ref{fig:ex_image_catdog} exemplarily shows attribution maps computed with different methods based on the VGG-16 model, for two object classes present in the ImageNet labels and the input image; ``Bernese Mountain Dog'' and ``Tiger Cat''.
Attributions in \mbox{Figs.~\ref{fig:ex_image_catdog}\emph{(a)}-\emph{(d)}} result from uniform rule application to the whole network.
Next to applications of \LRPZ~and \LRPALPHABETA~with \mbox{$\alpha=1$}, this includes Guided Backprop~\cite{springenberg2014striving} and Pattern Attribution~\cite{kindermans2017learning}.
Neither of these maps demonstrate class-discriminativeness and prominently attribute scores to the same areas, regardless of the target class chosen for attribution.
\LRPZ~additionally shows the effects of gradient shattering in a highly complex attribution structure due to its equivalence to G$\times$I.
Such attributions would be difficult to use and juxtapose in further algorithmic or manual analyses of model behavior.

To the right, attribution maps in Figs.~\ref{fig:ex_image_catdog}\emph{(e)}-\emph{(g)} correspond to variants of \LRPCMP,
which apply different decomposition rules
depending on layer type and position.
In Fig.~\ref{fig:ex_image_catdog}\emph{(e)}, the \mbox{\lrpflat-rule} is
not applied at all,
while in Fig.~\ref{fig:ex_image_catdog}\emph{(f)} it is used for the first three convolutional layers, and the whole convolutional stack --- including pooling layers --- in Fig.~\ref{fig:ex_image_catdog}\emph{(g)}.
Both heatmaps in Fig.~\ref{fig:ex_image_catdog}\emph{(e)} and Fig.~\ref{fig:ex_image_catdog}\emph{(f)} use $\alpha=1$.
Here altogether, the visualized attribution maps correspond more to an ``intuitive expectation'' of how relevance should be attributed  compared to Figs.~\ref{fig:ex_image_catdog}\emph{(a)}-\emph{(d)}, assuming a model predicts based on object understanding.
\mbox{Figs.~\ref{fig:ex_image_catdog}\emph{(e)}-\emph{(g)}} demonstrate the change in scale and semantic, from attributions to local features to a very coarse localization map, with changing placements of the \mbox{\lrpflat-rule}.
Further, it becomes clear that with an application of the \lrpalphabeta-rule in upper layers, object localization is lost (see  Fig.~\ref{fig:ex_image_catdog}\emph{(b)} vs. Fig.~\ref{fig:ex_image_catdog}\emph{(g)}),
while an application in lower layers avoids issues related to gradient shattering,
as shown in Figs.~\ref{fig:ex_image_catdog}\emph{(e)}-\emph{(f)} compared to~Fig.~\ref{fig:ex_image_catdog}\emph{(a)}.

Note that the special case shown in Fig.~\ref{fig:ex_image_catdog}\emph{(g)} is highly similar
to an application of the Class Activation Mapping (CAM)~\cite{zhou2016learning} method in the fully connected part of the model,
however replaces the upsampling over the model's convolutional stack of the CAM approach with the \lrpflat~decomposition based approach of the LRP framework, and is thus naturally capable of distributing negative relevance scores.

Note that the VGG-16 network used here never has been trained in a multi-label setting.
Despite only receiving one object category per input sample, it has learned to distinguish between different object types shown in the same image,
\eg that a dog is not a cat.
This in turn reflects well in the attribution maps computed after the \LRPCMP~pattern.

Further examples akin to Fig.~\ref{fig:ex_image_catdog} are given in the Appendix.


\subsection{Quantitative Results}
\label{sec:results-quantitative}
\begin{figure}[t]
  \centering
  \includegraphics[width=\linewidth]{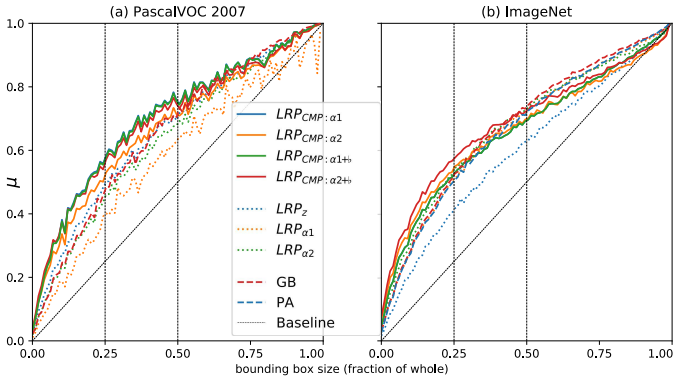} %
  \caption{
    Average in-total ratio $\mu$ as a function of bounding box size.
    Vertical lines mark thresholds of $25\%$ and $50\%$ covered image area.
    The baseline can be reached by uniformly attributing to all pixels of the image.
    Higher values are better.}
  \label{fig:itr_curves_joined}
\end{figure}

Figs.~\ref{fig:itr_curves_joined}\emph{(a)} and~\emph{(b)} show the average in-total ratio $\mu$ as a function of bounding box size, discretized over 100 equally spaced intervals, for PVOC 2007 and ImageNet.
Averages for $\mu$ and $\mu_w$ over the whole (and partial) datasets can be found in Tab.~\ref{tab:metrics_both_datasets}.
Large values indicate more precise attribution to the relevant object.

The inside-total relevance ratio highly depends on the size of the bounding box.
In addition to the average $\mu$ and $\mu_w$ as an aggregate over all classes and images, we also report $\mu_{\le0.25}$ and $\mu_{\le0.5}$, the average values over all objects whose bounding box does not span more than $0.25$ and $0.5$ times the area of the whole image respectively.
The assumed \emph{Baseline}
is the uniform attribution of relevance
over the whole image,
which is outperformed by all methods.
%

\begin{table}
    \caption{
        Average context attribution metrics for different analyzers and datasets.
        Row order is determined by $\mu_w$. Higher $\mu_\ast$ are better.
    }

    \begin{center}
    \begin{tabular}{ll|llll}

        Data & Analyzer & $\mu_w$  & $\mu_{\leq 0.25}$   &$\mu_{\leq 0.5}$ &  $\mu$ \\
        \hline
        \multirow{8}*{\rotatebox{90}{\parbox{3cm}{\centering PVOC\\(\emph{CaffeNet})}}}

        & \LRPCNNALPHATWOBETAONEFLAT    &              \textbf{2.716} &             \textbf{0.307} &                     0.421 &     0.532 \\
        & \LRPCNNALPHAONEBETAZERO       &                       2.664 &                      0.306 &            \textbf{0.426} &     \textbf{0.539} \\
        & \LRPCNNALPHAONEBETAZEROFLAT   &                       2.598 &                      0.301 &                     0.421 &     0.535 \\
        & \LRPCNNALPHATWOBETAONE        &                       2.475 &                      0.276 &                     0.388 &     0.504 \\
        & \LRPZ                         &                       2.128 &                      0.236 &                     0.353 &     0.480 \\
        & \textbf{\emph{GB}}                     &                       1.943 &                      0.212 &                     0.335 &     0.470 \\
        & \LRPALPHATWOBETAONE           &                       1.843 &                      0.205 &                     0.320 &     0.452 \\
        & \LRPALPHAONEBETAZERO          &                       1.486 &                      0.163 &                     0.273 &     0.403 \\
        & Baseline                      &                       1.000 &                      0.100 &                     0.186 &    0.322\\

        \hline
        \hline
        \multirow{10}*{\rotatebox{90}{\parbox{3cm}{\centering ImageNet\\(\emph{VGG-16})}}}

        & \LRPCNNALPHATWOBETAONEFLAT    &              \textbf{1.902} &             \textbf{0.397} &            \textbf{0.534} &     \textbf{0.714} \\
        & \LRPCNNALPHATWOBETAONE        &                       1.797 &                      0.368 &                     0.505 &     0.693 \\
        & \LRPCNNALPHAONEBETAZERO     &                  	1.7044 &                      0.3467 &                     0.4887 &     0.6898  \\
        & \LRPCNNALPHAONEBETAZEROFLAT &                  	1.7043 &                      0.3466 &                     0.4886 &     0.6898 \\
        & \LRPALPHATWOBETAONE                &                  1.702 &                      0.332 &                     0.496 &     0.706 \\
        & \textbf{\emph{GB}}                          &                  1.640 &                      0.312 &                     0.485 &     0.710 \\
        & \LRPALPHAONEBETAZERO               &                  1.609 &                      0.306 &                     0.475 &     0.699 \\
        & \textbf{\emph{PA}}                          &                  1.591 &                      0.303 &                     0.471 &     0.698 \\
        & \LRPZ                              &                  1.347 &                      0.236 &                     0.389 &     0.632 \\
        & Baseline                           &                  1.000 &                      0.128 &                     0.260 &     0.547\\

    \end{tabular}
    \end{center}
    \label{tab:metrics_both_datasets}
\end{table}

\LRPZ~performs noticeably worse on ImageNet than on PVOC,
which we trace back to the significant difference in model
depth (13 vs 21 layers)
affecting gradient shattering.
We omit \LRPEPS~in Tab.~\ref{tab:metrics_both_datasets} due to the identity in results to \LRPZ.
\LRPALPHABETA~has the tendency to attribute to all shown objects (via generally neuron-activating features) and suffers from the multiple object classes per image in PVOC,
where ImageNet shows only one class.
Also, the similarity of attributions between \emph{PA} and \LRPALPHABETA~with $\alpha=1$~observed in Fig.~\ref{fig:ex_image_catdog} seem consistent on ImageNet and result in close measurements in Tab.~\ref{tab:metrics_both_datasets}.

Tab.~\ref{tab:metrics_both_datasets} demonstrates that \LRPCMP~clearly outperforms
other methods consistently on large datasets.
That is, the increased precision in attribution to relevant objects is especially evident in the presence of smaller bounding boxes in $\mu_w$.
This can also be seen in $\mu_{\leq 0.25}$ and $\mu_{\leq 0.5}$ in Tab.~\ref{tab:metrics_both_datasets} and the left parts of Figs.~\ref{fig:itr_curves_joined}\emph{(a)} and~\emph{(b)}, where a majority of the image shows contextual information or other classes.
Once bounding boxes become (significantly) larger and cover over $50\%$ of the image, all methods converge towards perfect performance, as expected.
In both settings, \LRPCNNALPHATWOBETAONEFLAT~yields the best results,
while overall the composite strategy is more effectful than a fine tuning of decomposition rule parameters.

\subsection{Conclusion}
In this study, we discuss a recent development in the application of Layer-wise Relevance Propagation.
We summarize this emerging strategy of a composite application of multiple purposed decomposition rules as \LRPCMP~and
juxtapose its effects to previous approaches to LRP and other methods, which uniformly apply a single decomposition rule to all layers of the model.
For the first time, our results show that \LRPCMP~does not only yield \emph{measurably} more representative attribution maps,
but also provides a solution against gradient shattering affecting previous approaches,
and improves properties related to object localization and class discrimination via attribution.
Moreover, \LRPCMP~is
able to precisely attribute negative relevance scores to class-contradicting features while requiring only one modified backward pass though the model, using established tools from the LRP framework.
The discussed beneficial effects are demonstrated qualitatively and verified quantitatively at hand of two large and widely used computer vision datasets.

\bibliographystyle{IEEEbib}
\bibliography{references}



\newpage
\onecolumn
\section*{Appendix}
\label{sec:appendix}

\begin{figure*}[!h]
  \begin{center}
    \includegraphics[width=\linewidth]{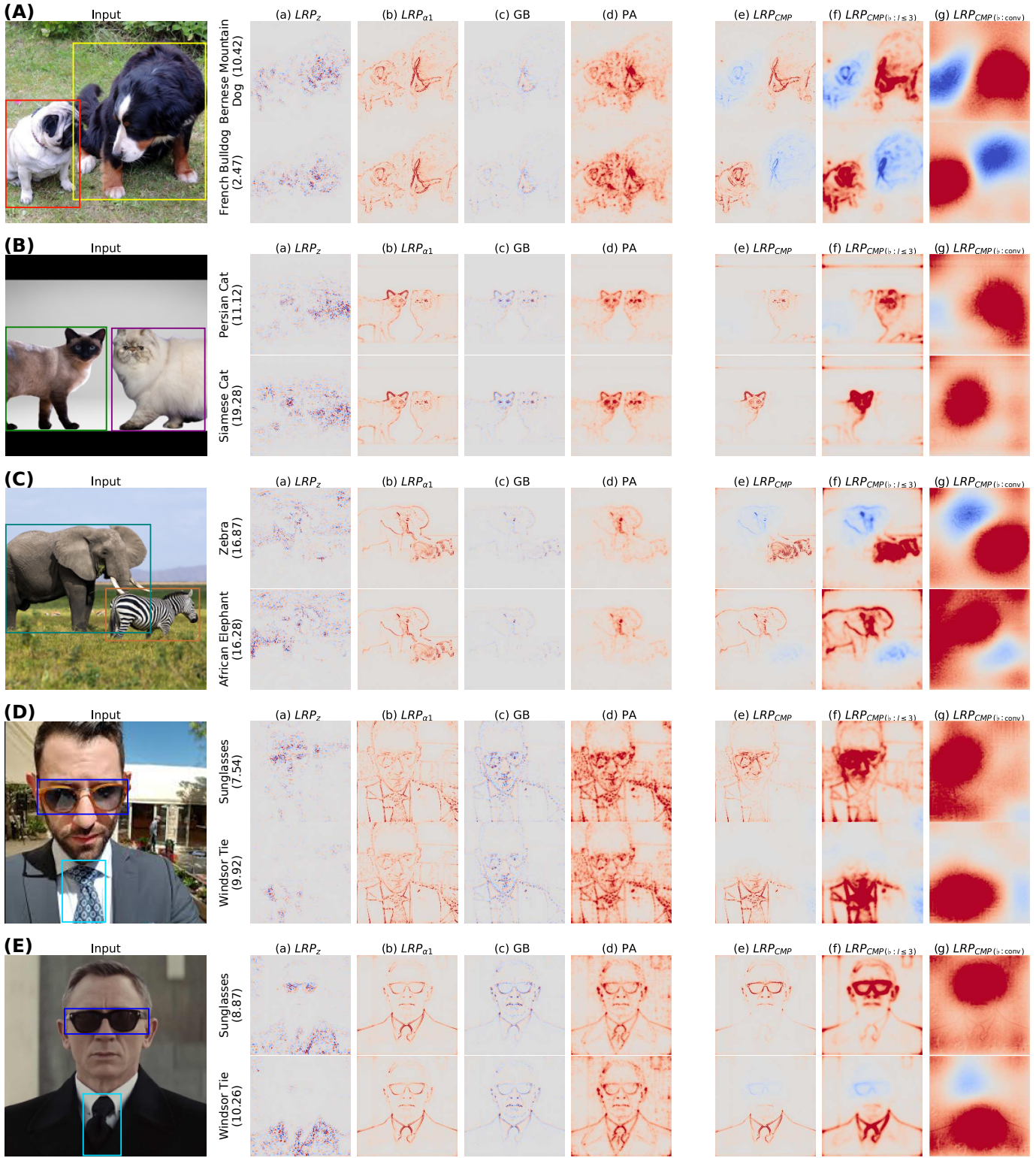}
    \caption{Different attributions for the output classes
    ``Bernese Mountain Dog'' and ``French Bulldog'' (A),
    ``Persian Cat'' and ``Siamese Cat'' (B),
    ``Zebra'' and ``African Elephant'' (C) and
    ``Sunglasses'' and ``Windsor Tie'' (D and E),
    using the pretrained VGG-16 model.
    For details \cf Fig.~\ref{fig:ex_image_catdog}.
    }
    \label{fig:ex_appendix}
  \end{center}
\end{figure*}

In Fig.~\ref{fig:ex_appendix} we provide further illustrative examples similar to Fig.~\ref{fig:ex_image_catdog},
using different input images and object classes.

\end{document}